\documentclass[sigconf]{acmart}

\usepackage{longtable}
\usepackage{tikz}
\usepackage{pgfplots}
\usepackage{csquotes}
\usepackage{algorithmic}
\usepackage{graphicx}
\usepackage{textcomp}
\usepackage{subcaption}
\usepackage{booktabs}
\usepackage{graphicx}
\usepackage{xcolor}
\usepackage{array}
\usepackage{soul}
\AtBeginDocument{%
  }

\newtheorem{definition}{Prompt}

\newcommand{\statedefsolid}[2][0.9\linewidth]{
  \par\noindent\tikzstyle{mybox} = [fill=yellow!20,
   thick,rectangle,inner sep=3pt,path picture={\fill [green!50!black] ([xshift=-8.15cm]path picture bounding box.north) rectangle (path picture bounding box.south west);}]
  \begin{tikzpicture}
   \node [mybox] (box){%
    \begin{minipage}{#1}{#2}\end{minipage}
   };
  \end{tikzpicture}
}

\copyrightyear{2024}
\acmYear{2024}
\setcopyright{acmlicensed}\acmConference[CIKM '24]{Proceedings of the 33rd
ACM International Conference on Information and Knowledge
Management}{October 21--25, 2024}{Boise, ID, USA}
\acmBooktitle{Proceedings of the 33rd ACM International Conference on
Information and Knowledge Management (CIKM '24), October 21--25, 2024,
Boise, ID, USA}
\acmDOI{10.1145/3627673.3680088}
\acmISBN{979-8-4007-0436-9/24/10}

\begin{document}



\title{iRAG: Advancing RAG for Videos with an Incremental Approach}



\author{Md Adnan Arefeen}
\authornote{Work was done while working as a research intern at NEC Laboratories America.}


\orcid{0000-0001-6486-8181}
\affiliation{%
  \institution{University Of Missouri-Kansas City}
  \city{Kansas City}
  \state{Missouri}
  \country{USA}
}
\email{aa4cy@mail.umkc.edu}


\author{Biplob Debnath}
\orcid{0009-0006-6932-0311}
\affiliation{%
  \institution{NEC Laboratories America}
  \city{Princeton}
  \state{New Jersey}
  \country{USA}
}
\email{biplob@nec-labs.com}

\author{Md Yusuf Sarwar Uddin}

\orcid{0000-0003-2184-0140}
\affiliation{%
  \institution{University Of Missouri-Kansas City}
  \city{Kansas City}
  \state{Missouri}
  \country{USA}
}
\email{muddin@umkc.edu}

\author{Srimat Chakradhar}
\orcid{0000-0003-3530-3901}
\affiliation{%
  \institution{NEC Laboratories America}
  \city{Princeton}
  \state{New Jersey}
  \country{USA}
}
\email{chak@nec-labs.com}


\begin{abstract}

Retrieval-augmented generation (RAG) systems combine the strengths of language generation and information retrieval to power many real-world applications like chatbots. Use of RAG for understanding of videos is appealing but there are two critical limitations. One-time, upfront conversion of all content in large corpus of videos into text descriptions entails high processing times. Also, not all information in the rich video data is typically captured in the text descriptions. Since user queries are not known apriori, developing a system for video to text conversion and interactive querying of video data is challenging. 

To address these limitations, we propose an incremental RAG system called {\it iRAG}, which augments RAG with a novel incremental workflow to enable interactive querying of a large corpus of videos. Unlike traditional RAG, iRAG quickly indexes large repositories of videos, and in the incremental workflow, it uses the index to opportunistically extract more details from select portions of the videos to retrieve context relevant to an interactive user query. Such an incremental workflow avoids long video to text conversion times, and overcomes information loss issues due to conversion of video to text,  by doing on-demand query-specific extraction of details in video data. This ensures high quality of responses to interactive user queries that are often not known {\it apriori}. To the best of our knowledge, iRAG is the first system to augment RAG with an incremental workflow to support efficient interactive querying of a large corpus of videos. Experimental results on real-world datasets demonstrate 23x to 25x faster video to text ingestion, while ensuring that latency and quality of responses to interactive user queries is comparable to responses from a traditional RAG where all video data is converted to text upfront before any user querying.

\end{abstract}






\maketitle

\section{Introduction}

Many applications routinely collect and store videos for offline analysis. For example, surveillance systems ensure public safety by monitoring public spaces, airports, transportation hubs, and critical infrastructure using video cameras~\cite{shibata2023listening}. These systems also store video feeds for offline analysis and investigations. Additionally, there is increasing usage of video analytics for patient monitoring in hospitals and healthcare facilities. Understanding information within videos is also essential for tasks like traffic monitoring, congestion management, and incident detection using feeds from traffic cameras~\cite{mittal2023ensemblenet}.

Large language models (LLMs) like ChatGPT~\cite{bubeck2023sparks} have demonstrated their proficiency in human-like text-based dialogues. Inspired by the success of LLMs in understanding extensive text repositories~\cite{arefeen2024leancontext}, recent proposals aim to extend this capability to video content analysis using LLMs~\cite{vlog, mmvid}. These approaches employ vision AI models to analyze each video clip, outputting the visual information as text descriptions. For instance, an object detector AI model can identify objects such as cars, trucks, bicycles, and persons in each frame, and also specify their locations in human-readable text. These AI models can process lengthy videos, capturing visual information as a text document that concatenates descriptions from each clip. LLMs can then utilize these text documents through advanced techniques like Retrieval-augmented Generation (RAG)~\cite{rag}, enabling them to provide insightful responses to interactive queries about the video content.

 Prior proposals to understand video content through interactive queries have two major limitations. First, the time required to process entire long videos and generate text descriptions by using heavyweight or complex AI models (which contain a large number of parameters and incur slow inference time) can be prohibitive. For example, it could take more than a day for heavyweight AI models to analyze visual information in a 24-hour surveillance video and generate text descriptions. This prevents the police from timely analysis of criminal incidents in the video. Second,  not all visual information in the videos is captured in the text descriptions, even if we use state of the art heavyweight vision AI models~\cite{carion2020end,taigman2014deepface,li2023blip,nguyen2022grit}. Furthermore,  the user queries are not known apriori, and it is difficult to determine which AI models should be used to convert visual content into text. 

To address the critical limitations of prior efforts to interactively query content in long videos, we propose {\it iRAG}. Unlike prior proposals~\cite{vlog, mmvid} that employ many AI models to extract textual information upfront, iRAG uses lightweight AI models (with fewer parameters and faster inference time) to quickly prepare 
an index for the content in the long videos. 
During preprocessing of the long video, our primary goal is to extract enough information from the video to be able to index the video data. Such an index facilitates effective extraction and retrieval of more detailed information from select sections of the video in response to interactive user queries. 

 While utilizing lightweight AI models reduces the indexing time, the text descriptions 
 may not be as rich, and many user queries may not be answerable by an LLM due to lack of facts in the text descriptions. To tackle this challenge, iRAG employs an incremental workflow. When the information extracted during indexing  is inadequate to successfully respond to a user query, additional AI models are used on-demand to extract more details from selective parts of the long video. This strategy ensures that iRAG extracts necessary detailed textual information on-demand, addressing specific queries as they arise, rather than collecting all textual information upfront without knowing the specific details needed to answer arbitrary user queries.

 iRAG has been deployed to process and analyze footage from the city-wide surveillance cameras. Since its deployment approximately one year ago, iRAG has significantly enhanced the capabilities of law enforcement agencies. By providing detailed insights, it has played a crucial role in investigating incidents, identifying suspects, and preventing potential crimes. iRAG's ability to swiftly and accurately analyze vast amounts of video data has made it an invaluable tool in maintaining public safety and security.

In summary, we make the following contributions: 
\begin{enumerate}

    \item We introduce a novel incremental RAG system called iRAG, which enables immediate interactive querying of large video repositories. Unlike traditional methods that require many AI models upfront in a lengthy preprocessing step to convert video into text before any user query can be answered, iRAG quickly indexes large videos using lightweight AI models and employs select heavyweight AI models in a query-aware manner whenever needed.

    \item  We propose a \textsf{Planner} in the iRAG system that leverages indexed information and the user query to quickly identify query-specific heavyweight AI models and the relevant sections of the video for detailed analysis. We also propose an \textsf{Extractor} that uses these AI models to convert information from the relevant video sections into text descriptions and update the index. Incremental extraction of content from video is useful for eliciting responses from an LLM.
    
    \item We evaluate iRAG on real-world datasets and demonstrate its capability to accelerate the video-to-text conversion process by a factor of 23-25x. Our novel incremental flow ensures that interactive user queries yield responses comparable in both latency and quality to those from a baseline traditional RAG system, where all video data is converted to text upfront before any user querying.

\end{enumerate}

\section{Incremental RAG}
\label{sec:irag}

\subsection{\bf Prior work to leverage context in video}

\begin{figure}[!htbp]
\includegraphics[width=0.9\linewidth]{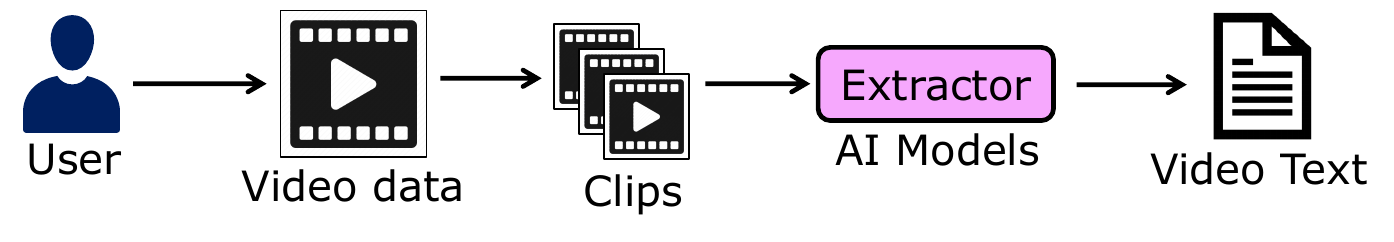}
        \caption{Video to text conversion}
     \label{fig:data_ingestion_rag}
\end{figure}

Prior work~\cite{vlog,mmvid} first produces a textual description of information in the entire video. The long video is divided into smaller clips. One or more generative AI models analyze these clips and produce textual descriptions for each. Concatenating these descriptions results in a long document. The entire preprocessing workflow is illustrated in Figure~\ref{fig:data_ingestion_rag}.

\begin{figure}[!htbp]
\centering
\includegraphics[width=\linewidth]{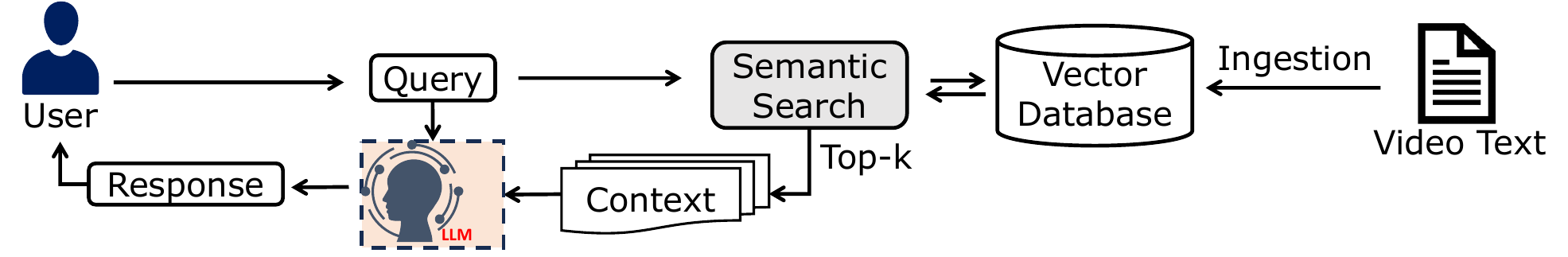}
        \caption{Conventional RAG workflow.}
     \label{fig:query-response}
\end{figure}

{\bf RAG:}
In Figure~\ref{fig:data_ingestion_rag}, a traditional Retrieval-augmented Generation (RAG)~\cite{rag} system is shown that augments the user prompt with query-specific context in the long document. The augmented prompt guides the LLM to produce more useful responses. We briefly review this process. The long text document is chopped up into {\it chunks} where a chunk can be a paragraph, or a sentence or some pre-determined number of consecutive characters. Each chunk is embedded into a vector of pre-determined length by an embedding algorithm, and these vectors are all stored in a vector database. All this makes it easy to identify chunks in the document that are similar to a user query.  For instance, utilizing the embedding model, we can embed a user query into a vector. Then, by computing the dot product of the query vector with each chunk vector in the database, top-$k$ chunks that are most similar to the user query are retrieved to form a query-specific context from the long document utilizing the semantic search block in Figure~\ref{fig:query-response}. 

Finally, the user query is augmented with query-specific context, and the augmented prompt results in a more contextually accurate response from an LLM.

\subsubsection{ \bf Limitations of Prior work}
We identify two critical limitations that make it difficult to use of prior work for advanced understanding of real-world videos. 

\textbf{Long preprocessing time:} Since user queries are not known apriori, many vision AI models are used to extract information from stored videos. However, models that capture more details are often computationally expensive. Preprocessing time is crucial, especially for time-sensitive tasks like crime analysis, where timely access to surveillance footage increases the chances of capturing critical evidence. As shown later, preprocessing a 24-hour surveillance video can take more than a day, delaying the use of this context in solving real-world problems and hindering law enforcement from gathering actionable intelligence, identifying suspects, and ensuring public safety.

\begin{figure}[!htpb]
  \centering
  \resizebox{0.9\columnwidth}{!}{
    \begin{tikzpicture} 
          \begin{axis}[
            ybar,                                       
            width=0.95\linewidth,                         
            height=4cm,                                 
            ylabel={Preprocessing Time (days)},                            
            ymin=0,                                     
            ymax=47,                                    
            symbolic x coords={1 day, 7 days, 30 days}, 
            xtick=data,                                 
            nodes near coords,                          
            nodes near coords align={vertical},         
            ]
            \addplot coordinates {(1 day, 1.57) (7 days, 11) (30 days, 46.4)};
          \end{axis}
    \end{tikzpicture}}
  \vspace{-4mm}
  \caption{Time to generate a long document from real-world videos}
  \vspace{-3mm}
  \label{fig:pre-ingestion-time}
\end{figure}

Figure~\ref{fig:pre-ingestion-time} shows the preprocessing bottleneck for querying and analyzing long videos from several New York City traffic intersections~\cite{piadyk2023streetaware}. We analyzed videos of 1, 7, and 30 days using the DETR object detection model~\cite{carion2020end} and the GRiT dense captioning model~\cite{nguyen2022grit}. On a server with an AMD Ryzen 5950X and NVIDIA GeForce RTX 3090, the average per-frame latency is about 70 ms for DETR and 1500 ms for GRiT. Processing one frame per second, the video-to-text generation for 1-day, 7-day, and 30-day videos takes approximately 1.57 days, 11 days, and 46.4 days, respectively. These lengthy conversion times hinder the development of practical systems for quickly querying and understanding long videos in real-world applications.

\textbf{Information loss:} During the conversion of video (vision domain) to text (text domain), there is a loss of information due to the shift in modalities. There are many details in the video that are not captured in the long text document, even if we use a lot of AI models during preprocessing. Since the user queries are not known apriori, often the long text document lacks sufficient information that can help the LLM provide a useful response to a user query. When this happens, there is no mechanism in prior approaches to go back to the stored videos and efficiently extract query-specific additional information (perhaps, by using new models that were not used or available during preprocessing).


\subsubsection{ \bf Our new approach} 
Instead of executing all vision AI models upfront to extract textual information from entire videos, we propose an {\it incremental} RAG (iRAG) system that does query-aware, on-demand extraction  of text data from select portions of the videos. 
Such a capability is not present in prior methods~\cite{vlog,mmvid} that solely rely on long preprocessing times to convert video to text {\it once} before any interactive querying of the long videos.

\subsection{Preprocessing in iRAG}
Unlike prior approaches, the goal of preprocessing in iRAG is to quickly build an index for the long video. For example, to answer the query \enquote{Is there a FedEx truck in a long video?}, the index (i.e. the object detection information) can be used to quickly identify the relevant clips where a truck was detected. Subsequent detailed extraction of text information from these selected clips (by using an AI model that can pick up text in images) can confirm if and where a FedEX truck is seen in the  long video. We remark that a variety of indexes can be quickly constructed by using lightweight vision AI models such as DETR~\cite{carion2020end}-an object detection model and CLIP~\cite{radford2021learning}, which can retrieve clips based on user query. The heavyweight vision AI models such as GRiT~\cite{nguyen2022grit} should be employed selectively, focusing only on video clips that align with the user's specified query. This targeted incremental approach ensures optimal use of computational resources while maintaining high accuracy in content analysis.

\subsection{Query-Response}
\statedefsolid{
\begin{definition}
\emph{You operate as a chatbot that is supported by a retrieval augmented generation system. You will utilize the
given context and your knowledge to
answer queries. If you are unable to
answer a query, your response is
``Unable to answer query. Please run additional models".}\\
Context: \{context\}\\
Query: \{query\}
\end{definition} 
}
Given a user query, iRAG retrieves the relevant context from the information collected during preprocessing by using the conventional RAG approach. Then, we construct a suitable prompt for the LLM by using the prompt template Prompt 1.
iRAG invokes the incremental flow when the LLM is unable to answer a user query. Next, we describe the iRAG system, and its key components.

\subsection{iRAG Components}
\begin{figure}[!tpb]
        \centering
        \includegraphics[width=\linewidth]{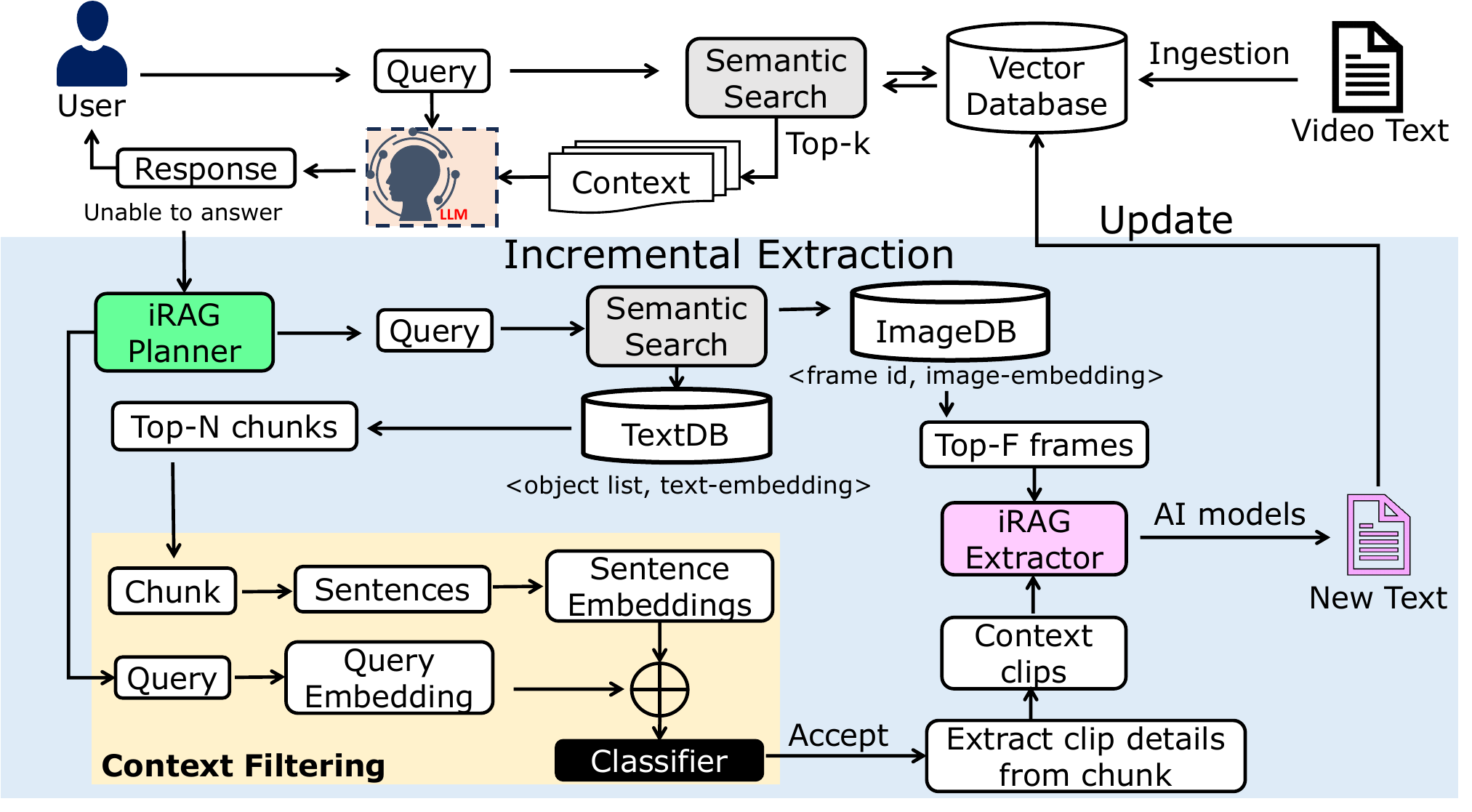}
        \vspace{-4mm}
        \caption{iRAG overview: The light blue rectangle denotes additional workflow compared to a conventional RAG.}
        \vspace{-3mm}
        \label{fig:qa_rag}
    \end{figure}

Figure~\ref{fig:qa_rag} provides an overview of the proposed iRAG system. The incremental workflow, which is activated only when the LLM cannot answer a query, is depicted within the light blue rectangle. There are two key components that support the query-aware, on-demand incremental flow:  \textsf{ planner}, and  \textsf{extractor}. 

The iRAG \textsf{Planner} retrieves query-related clips by leveraging indexed information and employs AI models via the \textsf{Extractor} to gather additional information needed for generating a response to the user query. This incremental process can loop up to a preset maximum, although typically, one iteration suffices to answer most queries. The query response includes a textual answer along with supporting video clips, aiding users in verifying the accuracy of the LLM's response. Future efforts may focus on integrating automatic response verifiers to ensure alignment between the textual response and video clips.

\subsubsection{\bf iRAG Planner}
\label{planner}
The workflow in our planner is demonstrated in Figure~\ref{fig:qa_rag}. Given a query, the \textsf{Planner} computes a query embedding vector by using the same embedding algorithm that the RAG system used to embed preprocessing text chunks into vectors in \textbf{TextDB} (generated from the output of the DETR model~\cite{carion2020end}), and retrieves the top-$N$ chunks in \textbf{TextDB} that are most similar to the query embedding vector. 

{\bf Improved context retrieval:} To enhance the quality of retrieved context, iRAG can maintain additional vector databases. For instance, iRAG can utilize \textbf{ImageDB} to store "frame vectors" related to clips. Lightweight AI models like CLIP~\cite{radford2021learning} embed visual information in each frame of the long video into a CLIP embedding vector, which serves as the frame vector. All frame vectors generated during preprocessing are stored in \textbf{ImageDB}. The \textsf{Planner} computes the CLIP embedding vector for the user query and retrieves the Top-$F$ frame vectors that are most similar to the user query vector. The frames corresponding to the Top-$F$ vectors are considered the Top-$F$ frames.
The context from the Planner comprises two components: the Top-$N$ context chunks from \textbf{TextDB} and the Top-$F$ context frames from \textbf{ImageDB}. iRAG then identifies related clips associated with these contexts. The iRAG \textsf{Planner} conducts further processing to reduce the top-$N$ chunks to a maximum of $k$ chunks, ensuring prompt and interactive query responses to the user.

{\bf Context Chunks Filtering}. After retrieving the context chunks, there may still be irrelevant clip information included. Extracting irrelevant video clip information at the incremental stage will add latency to the response. Therefore, the \textsf{Planner} ensures that we have at most $k$ important context clips associated to the chunks for the Extractor. A large $k$ implies a higher limit on the context clips, and the detailed extraction of so many clips will take time and adversely affect the quality of interactive querying. On the other hand, a very small $k$  can result in too few clips for detailed extraction, and the updated index may not have the context necessary for the LLM to provide a useful response to the user query. So, choice of $k$ is critical to ensure a balance between interactivity and usefulness of LLM responses.

{\bf Classifier:} iRAG employs a KNN classifier to group the top-$N$ chunks into two classes: ``accept" or ``reject".  The  embedding of each  top-$N$ check is concatenated with the query vector to form the feature vector for the classifier, which assigns an ``accept" or ``reject" label to each concatenated vector. Any vector having  ``reject" label is eliminated from further consideration. From the accepted chunks, the top $k$ chunks most similar to the query are selected for extraction. 

{\bf Training of KNN classifier:} We train a classifier by using data from our baseline system, and iRAG's \textsf{Planner}. Given a query, and a long video, we analyze the context retrieved by the baseline system, and iRAG. If a chunk retrieved by iRAG matches a chunk retrieved by the baseline system, then we assign a label of ``1" (accept); otherwise, we assign a label of ``0" (reject). 
Using the test queries in the VQA-2 dataset~\cite{vqav2}, we created labeled training data for our KNN classifier offline. 
After training, our KNN classifier assigns a label ``1" to an unlabeled iRAG context chunk if 3 out of 5 neighbors of the concatenated query-chunk vector  have the label ``1".

\subsubsection{\bf iRAG Extractor}
\label{extractor} 
The \textsf{Extractor} receives context clips from the \textsf{Planner}, along with a list of  heavyweight AI models to run on each clip. Instead of converting the entire video to text using a heavyweight model such as GRiT~\cite{nguyen2022grit}, the \textsf{Extractor} runs the heavyweight model only on the selected context clips deemed essential by the Planner for accurate query responses. The detailed text data extracted from these context clips is then used to update the \textbf{TextDB} vector database. Subsequently, the user query is re-issued with the updated context to generate a prompt for the LLM to obtain the response. Figure~\ref{fig:qa_rag} illustrates how the \textsf{Planner} and \textsf{Extractor} collaborate in the iRAG system to incrementally update the \textbf{TextDB}.

\section{Experimental results}
\label{sec:experiments}

We compare iRAG with the baseline video RAG system described in Section~\ref{sec:irag}. Baseline uses GRiT~\cite{nguyen2022grit} for preprocessing, while iRAG uses DETR~\cite{carion2020end} and CLIP~\cite{radford2021learning}. iRAG employs GRiT when  preprocessed information cannot answer a user query. We compare the pre-processing time, average query processing time, and accuracy of the two systems. We briefly describe our video datasets in Table~\ref{tab:dataset}, and the queries we pose on these datasets. We use outputs from the baseline system as the ground truth. 
\vspace{-3mm}

\subsection{Datasets}
\textbf{VQA-v2} validation dataset~\cite{vqav2}: This dataset has 40,504 images. Of the 214,354 queries that are posed on these images, we selected 7799 queries, all of which start with ``Is there". We concatenate the images corresponding to these queries to create a long video (about 2 hours and 10 minutes long at 1 frame per second). We also randomly choose 1000 queries from the 7799 queries. These 1000 queries are used to evaluate iRAG and the baseline system.

\textbf{MSR-VTT}~\cite{msrvtt}: This dataset has small clips of YouTube videos, with a caption per video. We concatenate several video clips to create a long video (about three and half hours long at 20fps). This dataset also has 318 queries.

\textbf{StreetAware:}~\cite{piadyk2023streetaware}: This dataset has 7.75 hours of video (2 TB of raw data) of traffic activity at four New York City traffic intersections. We selected a 46-minute video for our evaluation. This dataset does not have queries. As described later, we generated appropriate queries by suitably prompting an LLM.

\textbf{Tokyo MODI}~\cite{kossmann2023extract}:  This dataset is a video of a busy street in Tokyo. Like others, we also used 2 hours of this video~\cite{kossmann2023extract}. This dataset has no queries, and we prompt an LLM to generate queries for our evaluation.  

\begin{table}[!htpb]
\scriptsize
    \centering
    \vspace{-2mm}
    \caption{Datasets used for evaluating iRAG}
    \vspace{-3mm}
    \resizebox{0.9\columnwidth}{!}{
    \begin{tabular}{ccccc}
    \toprule
       Video Dataset & Type  & Duration & Keyframes & \# of test queries  \\
    \toprule
        VQA-v2 & Traditional & 02:09:59 & 7799 & 1000\\
    \midrule
        MSRVTT & Traditional & 03:29:35 & 6291 & 318\\
    \midrule
    StreetAware & Real-world & 00:46:45 & 2363 & 40\\
    \midrule
    Tokyo MODI & Real-world & 02:00:23  & 1444 & 14\\
    \bottomrule
    \end{tabular}}
    \label{tab:dataset}
\end{table}

\subsection{Queries}

Although two of our datasets (VQA-v2 and MSRVTT) come with queries, the real world datasets like StreetAware~\cite{piadyk2023streetaware} and MODI~\cite{kossmann2023extract} do not. Therefore, we use an LLM to generate appropriate queries. For this purpose, we ran the GRiT model~\cite{nguyen2022grit} on both videos to generate captions, which are then used as context for the following prompt:

\statedefsolid{
\begin{definition}
\emph{Given a context, generate a question for the context.
    Here are a few examples of questions: \\
    Return the question without any additional text. Think you are an investigator querying a textual description of a video.}\\
Context: \{context\}
\end{definition} 
}

By repeatedly prompting the LLM with various contexts, we generated 40 queries for the StreetAware dataset and 16 queries for the Tokyo MODI datasets.

\subsection{Implementation details}

We implemented iRAG using the Langchain framework~\cite{langchain}. The vector database (with the FAISS index~\cite{faiss}) is from Langchain, and the LLM is OpenAI's \texttt{gpt-3.5-turbo} model~\cite{openai} API. 
We conducted experiments on a server equipped with an AMD Ryzen 5950X 16-Core processor and an NVIDIA GeForce RTX 3090 GPU. For the classifier training, we select 1000 query from the training data of VQA-v2, for each query we run the GRiT model to extract ground truth clips. Then we label the each concatenated query-chunk embedding vector as 1 if the chunk contains clips from ground truth, other wise give a 0. Then we train KNN classifier on this. After the KNN is trained, we test it on 1000 queries from the validation samples from VQA-v2 dataset to report results.

\subsection{Evaluation metric}

For the baseline and iRAG, we retrieve $k$ chunks as context 
that correspond to video clips. Given a query, we compute the ratio of the number of video clips common to both systems and the number of video clips in the baseline. If the ratio is 1, then both iRAG and the baseline retrieve the same video clips as context for the query. We define {\it recall@k} as the average of these ratios for all queries in a dataset. By varying $k$, we can compute different recall@k and study the impact of $k$ on the quality of context retrieved by iRAG (compared to the baseline). Higher the value for recall@k, the closer are the responses from the LLM for the baseline and iRAG. However, naively increasing $k$ to improve context quality will also lead to an increased processing time in the incremental flow. Increasing $k$ can increase the number of video clips that will require detailed extraction during the incremental flow. This additional processing time can adversely affect interactive querying experience.   

\subsection{System evaluation}
We systematically evaluate the benefits from different enhancements in iRAG i.e. preprocessing time, benefit of chunk filtering, query processing time, impact of varying $k$ on the context quality and interactivity compared to the baseline.

\subsubsection{\bf Preprocessing time:} Preprocessing of the long videos using DETR is a lot faster than preprocessing with GRiT. Table~\ref{tab:ingestion_vs_acc} shows that iRAG's preprocessing time is a 23x to 25x less than the preprocessing time of the baseline.

\begin{table}[!htpb]
\scriptsize
\vspace{-2mm}
\caption{Preprocessing time of iRAG and baseline}
\vspace{-3mm}
    \centering
    \begin{tabular}{ccccl}
    \toprule
        Data  & System & Model & Preprocessing Time \\
       \midrule
       VQA-v2 & iRAG & DETR + CLIP & 48 min 29 sec \\
       & RAG & GRiT & 1093 min 36 sec \\\midrule
       MSRVTT  & iRAG & DETR + CLIP& 8 min 23 sec \\
        & RAG & GRiT & 199 min 50 sec \\

       \bottomrule
    \end{tabular}
    
    \label{tab:ingestion_vs_acc}
\end{table}

\subsubsection{\bf Benefit from using ImageDB:} 
By using lightweight vision-language models like CLIP, we can extract richer indexes in the preprocessing phase of iRAG.    
As shown in Table~\ref{tab:vqav2-clip}, the context video clips retrieved by iRAG when both CLIP and DETR models are used in preprocessing have a better context quality than the case where only DETR is used. Table~\ref{tab:vqav2-clip} shows the context quality recall@k for different values of $k$. For example, for the VQA dataset, when iRAG uses only DETR object detector for indexing (i.e. preprocessing), the recall@k for $k=20$ is $0.815$. This means that across all the queries in the dataset, there was on average about 81\% overlap of video clips retrieved as context by iRAG and the baseline.

Performance of iRAG on the StreetAware and MODI datasets are shown in Figure~\ref{fig:streetaware_examples} and Table~\ref{tab:MODI}, respectively. For these datasets, we show the human-understandable response video clips, and LLM responses from the iRAG system.

\begin{table}[!htpb]
    \centering
    \vspace{-2mm}
    \caption{Performance of iRAG on VQA-v2 and MSRVTT datasets.}
    \vspace{-3mm}
    \resizebox{0.9\columnwidth}{!}{\begin{tabular}{cccccccc}
    \toprule
    Dataset & System &  Models  & r@1 & r@2 & r@4 & r@8 & r@20 \\
     \midrule
     VQA-v2 & iRAG     &     DETR           & 0.2564 & 0.3593  & 0.4961 & 0.6368 & 0.8153\\
    &            &     DETR + CLIP           & \textbf{0.3299} & \textbf{0.4127}  & \textbf{0.5314} & \textbf{0.6607} &\textbf{0.8251}\\\midrule
    MSRVTT & iRAG & DETR & 0.4098 & 0.5135 & 0.6652 & 0.8003 & 0.9223\\
    & & DETR + CLIP & \textbf{0.4127} & \textbf{0.5206} & \textbf{0.6716} & \textbf{0.8028} & \textbf{0.9240}\\
    \bottomrule
    \end{tabular}}
    
    \label{tab:vqav2-clip}
\end{table}
\vspace{-3mm}

\subsubsection{\bf Benefit from chunk filtering:}
Results in Table~\ref{tab:vqav2-clip} do not consider the case where the chunks retrieved by the Planner module are filtered to eliminate the less important chunks. The benefit from filtering context chunks is shown in Table~\ref{tab:refine_benifit} and Table~\ref{tab:acc_vs_cost}. For the VQA-v2 dataset, with $k=8$, the Planner module retrieves 8 context chunks and the recall score is 66.07\%. However, with chunk filtering, based on the query, the filtering classifier is able to eliminate less important chunks from the proposed 50 chunks from the Planner. Thus, the average number of chunks selected by the Planner after filtering is only 6.88 while the recall score is 74.14\%. Across all the 1000 queries for this data set, the average number of clips sent for detailed extraction in the incremental flow was only 6.88 chunks rather than 8 chunks. This results in reduced processing times for detailed extraction with accuracy improvement for the iRAG system with the filtering module. Therefore, iRAG suggests a cost-effective approach to extracting data for a query.

\begin{table}[!htpb]
\scriptsize
    \centering
    \vspace{-2mm}
    \caption{Benefit from filtering context clips (VQA-v2 dataset)}
    \vspace{-3mm}
    \resizebox{\columnwidth}{!}{%
    \begin{tabular}{cccc}
    \toprule
       Method  & initial k &avg. \# of chunks per query & recall score \\
       \midrule
       iRAG [w/o filtering]  & 8 & 8 & 0.6607 \\
       \midrule
       iRAG [w filtering] &  50 & \textbf{6.88} & \textbf{0.7414}\\
       \bottomrule
    \end{tabular}}
    
    \label{tab:refine_benifit}
\end{table}
\vspace{-3mm}

As the increase of k indicates the increase in total query processing time due to extracting unnecessary details using the Extractor, we define this k as a cost metric. We observe that for a specific k value (cost), the iRAG with filtering offers improved accuracy (recall@k) score compared to iRAG w/o filtering. To visualize this, for iRAG, in Table~\ref{tab:acc_vs_cost}, we set the initial k that the query planner offers are 10-50 chunk proposals. If we follow the context clips by Planner, we may result in extracting frames from 10-50 chunks by the Extractor. Our iRAG with filtering refines the k value and select it on average from $1.49\sim6.88$ chunks. Now, if we observe the recall@k score for similar value of k ranging from 2-8 in iRAG w/o filtering, we observe that the iRAG with filtering improves the recall score by $1.9\% \sim 13.8\%$. Hence, the iRAG with filtering module is cost-effective and efficient.

\begin{table}[!htbp]
\scriptsize
\centering
\vspace{-2mm}
\caption{Impact of filtering on context quality}
\vspace{-3mm}
\label{tab:acc_vs_cost}
\resizebox{0.9\columnwidth}{!}{%
\begin{tabular}{cc|ccc}
\toprule
\textbf{iRAG w/o filtering}      &          &           &   \textbf{iRAG with filtering}        &     \\ \midrule
cost & recall@k & cost (k) &initial k &  recall@k       \\ 
(k) &  & [after filtering] &[before filtering] &        \\ \midrule
2        & 0.367    & \textbf{1.49} & 10 & \textbf{0.386} \\
         &          & \textbf{2.84} & 20 &\textbf{0.528} \\\midrule
4        & 0.501    & \textbf{4.17} & 30 & \textbf{0.639} \\\midrule
6        & 0.567    & \textbf{5.49} & 40 &\textbf{0.683} \\\midrule
8         & 0.624         & \textbf{6.88} & 50 &\textbf{0.741} \\ \bottomrule
\end{tabular}%
}
\end{table}
\vspace{-3mm}

\subsubsection{\bf Query processing time vs $k$:}
\begin{figure}[!htpb]
    \centering
    \includegraphics[width=0.8\linewidth]{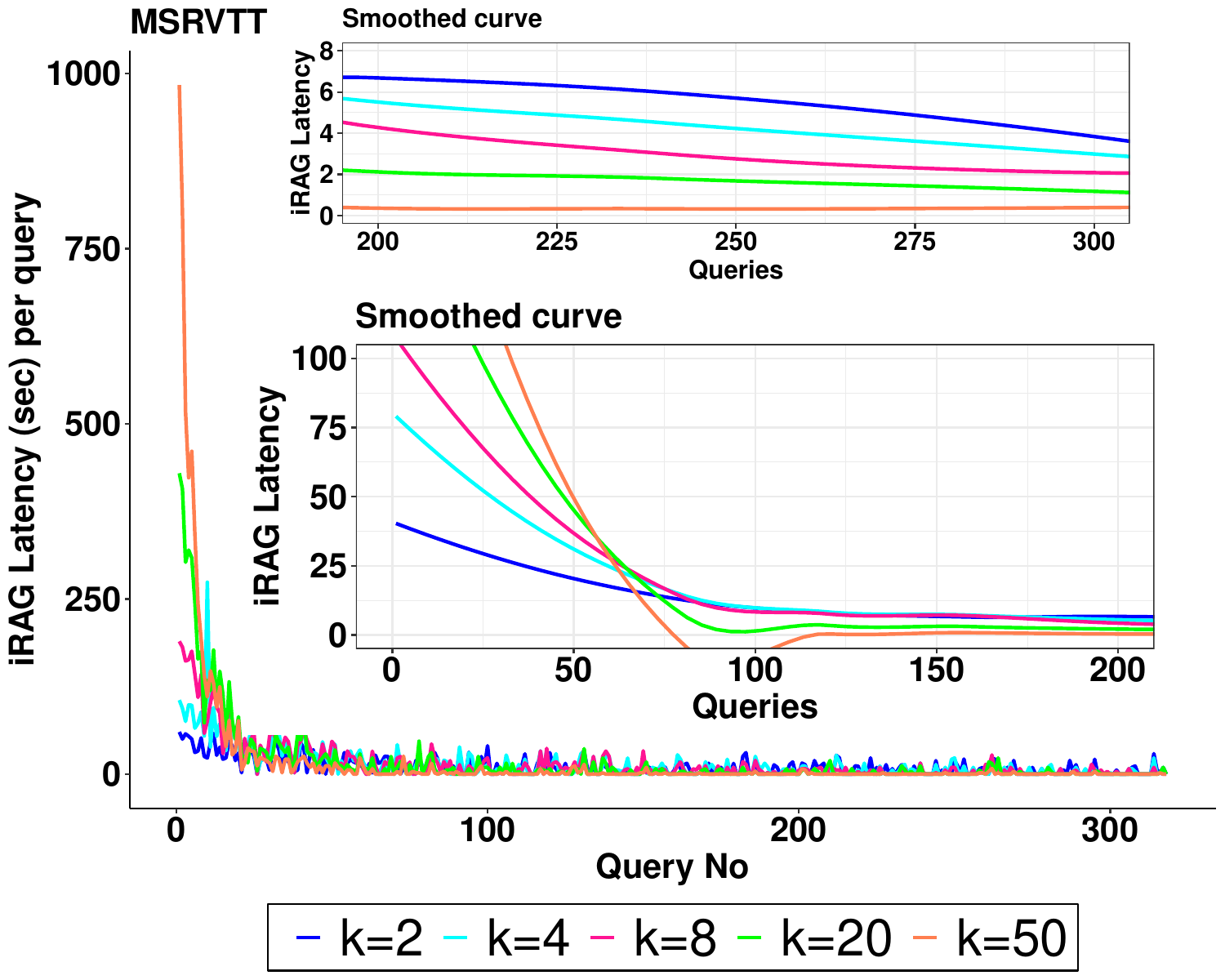}
    \vspace{-2mm}
    \caption{Query processing time distribution for different values of $k$}
    \vspace{-3mm}
    \label{fig:irag-exec-time}
\end{figure}

Query processing time is important. It determines if the query-response session is interactive. It includes retrieval of the context clips by the Planner optimized by filtering, and detailed extraction of video clips by the Extractor. Figure~\ref{fig:irag-exec-time} shows how interactivity (the query processing time) varies with $k$. 

There are two important characteristics of iRAG. First, from the inset that shows the query processing time for the first 200 queries, we observe that the query processing time for the first few queries is quite high. However,  the query processing time drops dramatically as we process more queries. This is because the information from the detailed extraction of clips for the first few queries is added to the text and image vector DBs, and subsequent queries do not trigger a lot of detailed extraction. Second, as we increase $k$, the number of context clips increase and the incremental extraction time increases, leading to an increase in query processing time. For example, the query processing time for $k=2$ for the first few queries is approximately 25-30 seconds. However, for the same queries, the query processing time jumps to 70-75 seconds when $k=4$. However, irrespective of the value of $k$, as shown in the inset for queries 200 - 1000, query processing time drops rapidly to less than 5 seconds as we process more queries. We also re-ordered the 1000 queries a few times, and observed that the above observations hold. 

Filtering context clips by the Planner eliminates less important clips, and improves interactivity. This is because due to filtering, not all the $k$ context clips are sent for detailed extraction. Thus filtering context clips effectively reduces $k$ for the Extractor module, and the reduced detailed extraction processing time leads to  significantly improved interactivity. Note that the LLM query execution time remains almost constant across queries for both iRAG and the baseline. As a result, we have excluded this time from the query processing time calculations.

\subsubsection{ \bf Incremental extraction:}
The baseline extracts detailed information of all the video clips during preprocessing. In contrast, iRAG does opportunistic, query-driven detailed extraction of clips as necessary.  Table~\ref{tab:fraction-video} shows the fraction of the clips for which the Extractor performed detailed extraction for different values of $k$. For example, after processing all the queries in MSRVTT dataset, and even for large values of $k$, the Extractor performed detailed extraction for only about half the video clips. 

\begin{table}[!htpb]
\centering
\vspace{-2mm}
\caption{Detailed extraction statistics}
\vspace{-3mm}
\label{tab:fraction-video}
\resizebox{0.9\columnwidth}{!}{%
\begin{tabular}{@{}ccc|ccc@{}}
\toprule
k &
  \begin{tabular}[c]{@{}c@{}}Fraction of\\ video extracted \\ (VQA-v2)\end{tabular} &
  \begin{tabular}[c]{@{}c@{}}Fraction of\\ video extracted \\ (MSRVTT)\end{tabular}  & k &
  \begin{tabular}[c]{@{}c@{}}Fraction of\\ video extracted \\ (VQA-v2)\end{tabular} &
  \begin{tabular}[c]{@{}c@{}}Fraction of\\ video extracted \\ (MSRVTT)\end{tabular}\\ \midrule
2  & 0.437 & 0.323 & 10 & 0.771 & 0.458 \\\midrule
4  & 0.578 & 0.392 &20 & 0.907 & 0.493 \\\midrule
6  & 0.670 & 0.428 &30 & 0.968 & 0.503 \\\midrule
8  & 0.724 & 0.445 & 40 & 0.986 & 0.508 \\\midrule
\end{tabular}%
}
\end{table}
\vspace{-3mm}

For VQA dataset, iRAG processes about 77\% of the images for reasonable values of $k$ (between 2 and 10). The percentage is a higher because the long video was artificially created by us. Here, several images are packed into a video clip, and we do detailed extraction on clips rather than images. If we treat a video clip to be a frame, then the fraction of images processed by the Extractor is similar to the MSRVTT dataset case.

\subsubsection{\bf Human Evaluation}

\begin{figure}[!htpb]
    \centering
    \includegraphics[width=0.8\linewidth]{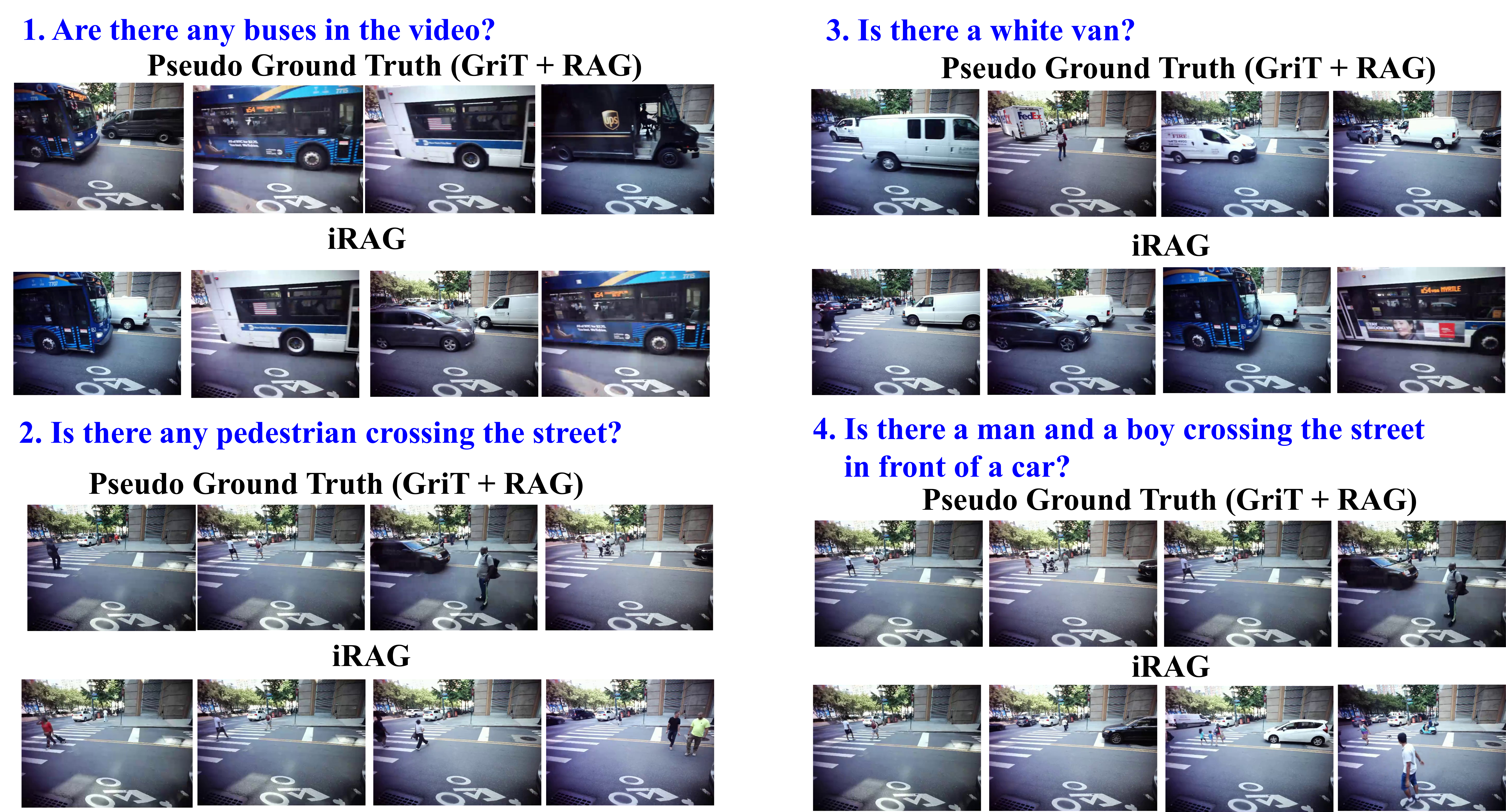}
    \vspace{-3mm}
    \caption{iRAG's qualitative evaluation on StreetAware videos~\cite{piadyk2023streetaware}}
    \label{fig:streetaware_examples}
    \vspace{-2mm}
\end{figure}

Figure~\ref{fig:streetaware_examples} shows frames in the context clips retrieved by iRAG and the baseline for a variety of queries. Visual inspection also shows that the context clips retrieved in iRAG is similar to the context clips retrieved by the baseline.

Table~\ref{tab:MODI} shows the LLM responses to queries in the Tokyo MODI dataset. In almost all cases, the responses from iRAG align well with the responses from the baseline system. We also highlight a case in which the user query is ambiguous, specifically asking about the color of a car on the street. Both the baseline and our approach successfully retrieve a car, but they differ in the color identified.

\begin{table}[!htpb]
\scriptsize
\centering
\caption{LLM response on MODI dataset~\cite{kossmann2023extract}}
\vspace{-3mm}
\label{tab:MODI}
{\tiny\renewcommand{\arraystretch}{0.3}
\resizebox{!}{0.13\paperheight}{%
\begin{tabular}{p{0.15\linewidth} | p{0.1\linewidth}| p{0.1\linewidth} | p{0.25\linewidth} }


\toprule
\textbf{Query} &
  \textbf{Baseline} &
  \textbf{iRAG} & \textbf{Comments} \\
  \midrule

Were there any black cars on the road? &
Yes& 
Yes&

There was a black car on the road at the time at 1:52:10 and 1:36:05 \\

  \midrule
  
Were there any people riding bicycles? &
Yes&
Yes&
There was a person riding a bicycle at 1:12:50, 1:37:15 and 1:58:15.\\
 
  \midrule
Were there any green trees lining the sidewalk? &
Yes&
Yes&
There were green trees lining the sidewalk at  0:10:35, 0:29:45, 0:42:20, 1:11:15, 1:35:20, 1:52:10, and 1:55:10.\\
  
  \midrule
Were there any people wearing hats? &
Yes &
Yes &
There were people wearing hats at 1:19:50, and 1:57:55.\\
  
  \midrule

Were there any green trees with lots of leaves? &
Yes &
Yes &
There are multiple instances of green trees with lots of leaves mentioned at 0:19:35 and 0:15:35\\

  \midrule

Were there any people wearing checkered shirts? &

  I can not answer. &
  
  I can not answer. &
  Based on the given context, there were no mentions of people wearing checkered shirts\\
  \midrule

Were there any people wearing white jackets? &
Yes &
Yes &
There is evidence of a person wearing a white jacket in the following segment: 0:43:05, 1:45:35, and 1:54:30.\\

  \midrule
  
What color is the car on the street? &
\sethlcolor{yellow}\hl{Green and white}&
\sethlcolor{yellow}\hl{Black} &
In this case, both answers are correct as the time a black car is seen at 1:53:10 and green and white cars are seen at 1:28:15.\\

  \bottomrule
\end{tabular}
}
}
\end{table}
\vspace{-4mm}

\section{Prototype Implementation}
\begin{figure}[!tbp]
    \centering
    \includegraphics[width=0.9\linewidth]{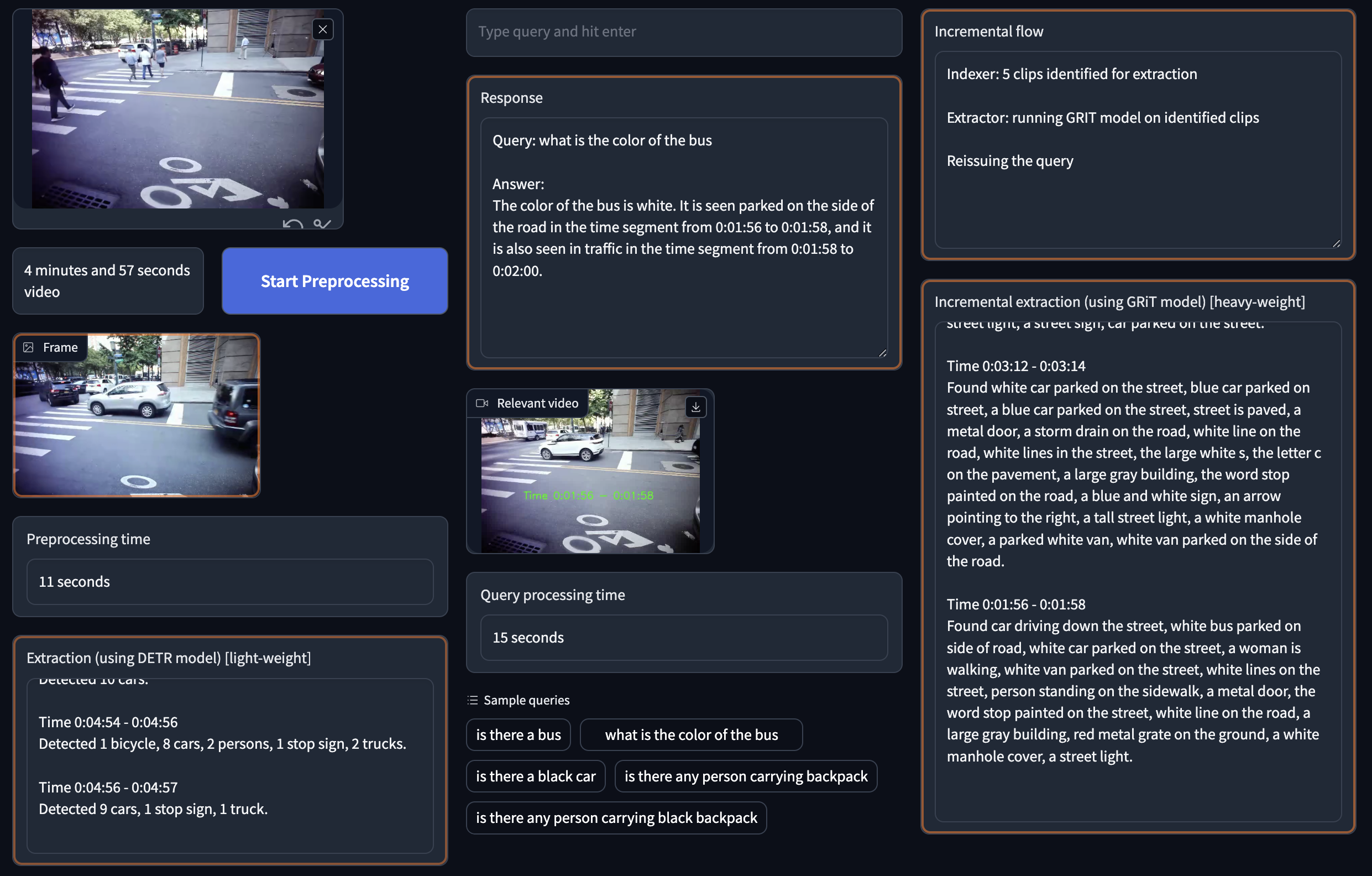}
    \caption{iRAG prototype implementation}
    \vspace{-4mm}
    \label{fig:prototype}
\end{figure}

We have developed a prototype that allows users to interact with the iRAG system. Figure~\ref{fig:prototype} illustrates the user interface for the iRAG system. Initially, the user uploads a video and starts the preprocessing stage by clicking on ``Start Preprocessing," utilizing fast AI models. Once preprocessing is complete, the iRAG system is ready to process the query.

In this example, the DETR object detection model~\cite{carion2020end} is employed for preprocessing. Now, for detection-based queries such as ``Is there a bus?", the indexer efficiently indexes the video and provides the query's answer. For more advanced user queries, like ``What is the color of the bus?", the indexer suggests the necessary video clips. The Extractor then uses the GRiT model~\cite{nguyen2022grit} to extract detailed text information from these clips. Subsequently, the extracted text serves as context, and the LLM API is invoked along with the query to generate the final response.

\section{Related Work}
Traditional video understanding models have primarily concentrated on designing task-specific models, incorporating various neural network architectures and training techniques, and excelling in the analysis of short video clips, as evidenced by previous works~\cite{TSM, slowfast, timesformer, vidtr, vivit, uniformer, cosine_sim_icpr}. However, this paper shifts its focus to analyzing long videos.

In recent years, significant progress has been made in multimodal learning, particularly in integrating video and language understanding, as demonstrated by recent research~\cite{li2023lavender, li2023videochat, maaz2023video}. Researchers have explored techniques like ``visual instruction tuning'', which combines a fixed Language Model (LLM) with a vision model and adaptable modules to refine LLMs' ability to generate textual descriptions for video content~\cite{liu2023visual, zhang2023llavar}. While these models are utilized in systems for answering questions about videos~\cite{wang2023vqa, hu2023promptcap, zhang2023pmc,arefeen2024vita}, most of them are tailored for short video clips or images. This paper, however, aims to improve the runtime efficiency of the video-to-text extraction process specifically for long videos within a retrieval-augmented generation (RAG)~\cite{rag} based system.

Video question answering systems~\cite{wang2023vqa,hu2023promptcap,zhang2023pmc} have been extensively researched, but primarily on short video clips or images. Recent works, however, have shifted focus to understanding complex relationships in long videos, such as cause-and-effect and sequential order~\cite{gao2021env,gao2023mist}. Long-form video question-answering systems typically involve reasoning about multiple events, various levels of detail, and causality. Efficiently segmenting videos for each query can be repetitive and challenging, especially for long videos.

\section{Conclusion}
We introduced an innovative incremental Retrieval-augmented Generation (iRAG) system to enhance the understanding of videos. Instead of using many AI models to extract text data from the entire video corpus upfront, iRAG quickly indexes the video data using lightweight AI models. When the extracted information during indexing is insufficient for certain user queries, iRAG employs an incremental workflow to extract more detailed information from select, query-relevant video data using heavyweight AI models. This new information is then used to retrieve an updated context, resulting in a useful response from the LLM to user query. While we evaluated iRAG for interactive video understanding, it is applicable to a variety of other non-textual data types like audio, LiDAR, and time-series data. The required AI models for such data differ, but iRAG remains indispensable due to its ability to handle time-sensitive non-textual content efficiently.

\bibliographystyle{ACM-Reference-Format}
\bibliography{sample-base}

\end{document}